\PassOptionsToPackage{table}{xcolor}
\documentclass[onecolumn]{openhelix}
\usepackage[most]{tcolorbox}

\usepackage{amsmath}
\usepackage{amssymb}
\usepackage{amsfonts}
\usepackage{tabularx}
\usepackage{float}
\usepackage{array}
\usepackage{longtable}
\usepackage{booktabs}
\usepackage{makecell}
\usepackage{xspace}
\usepackage{mathtools}
\usepackage{amsthm}
\usepackage{multirow}
\usepackage{graphicx}
\usepackage{nicefrac}
\usepackage{microtype}
\usepackage{afterpage}
\usepackage{pifont}

\usepackage{wrapfig}
\usepackage{enumitem}

\newcommand{\huggingface}{\raisebox{-1.5pt}{\includegraphics[height=1.05em]{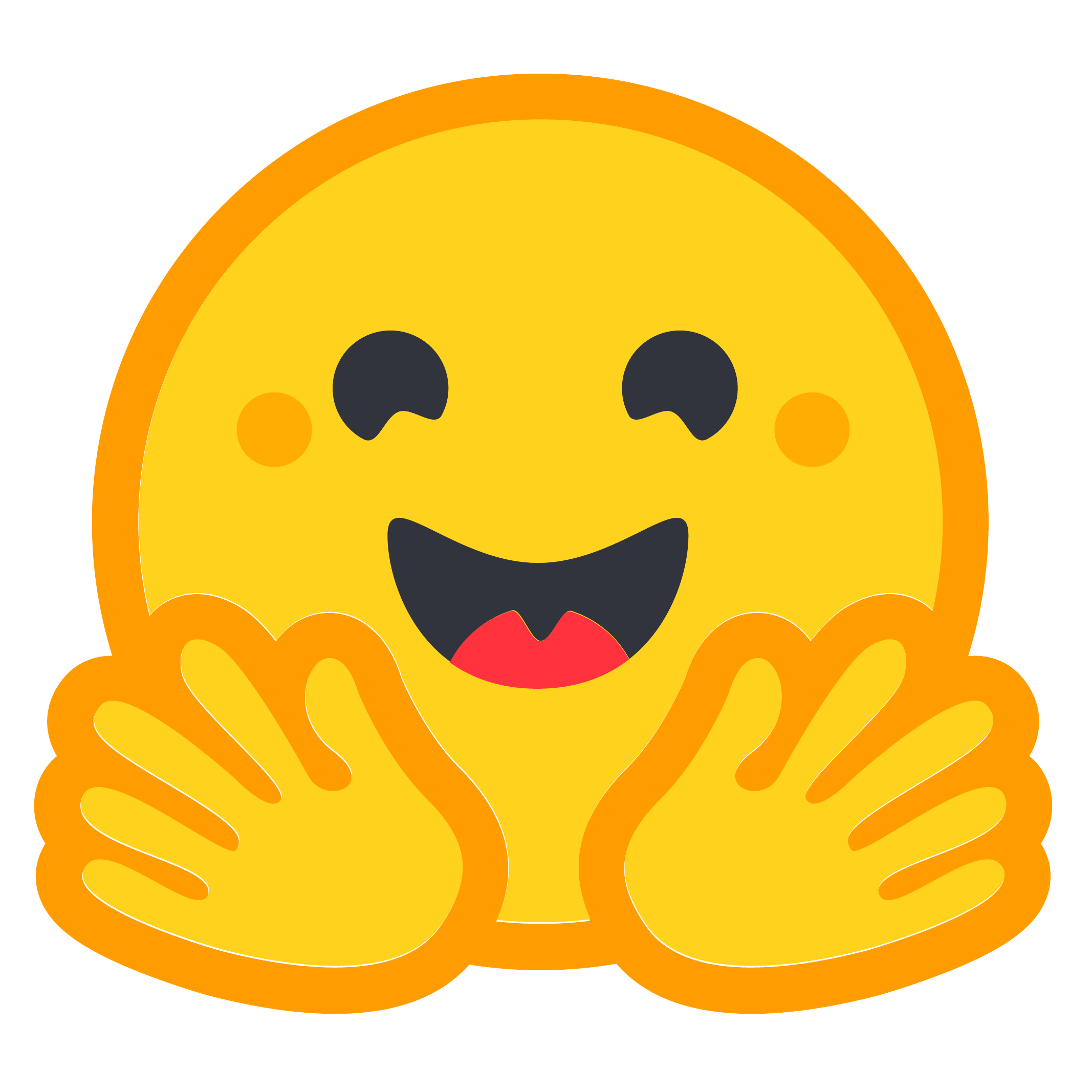}}\xspace}
\newcommand{\hfdataset}{\raisebox{-1.5pt}{\includegraphics[height=1.05em]{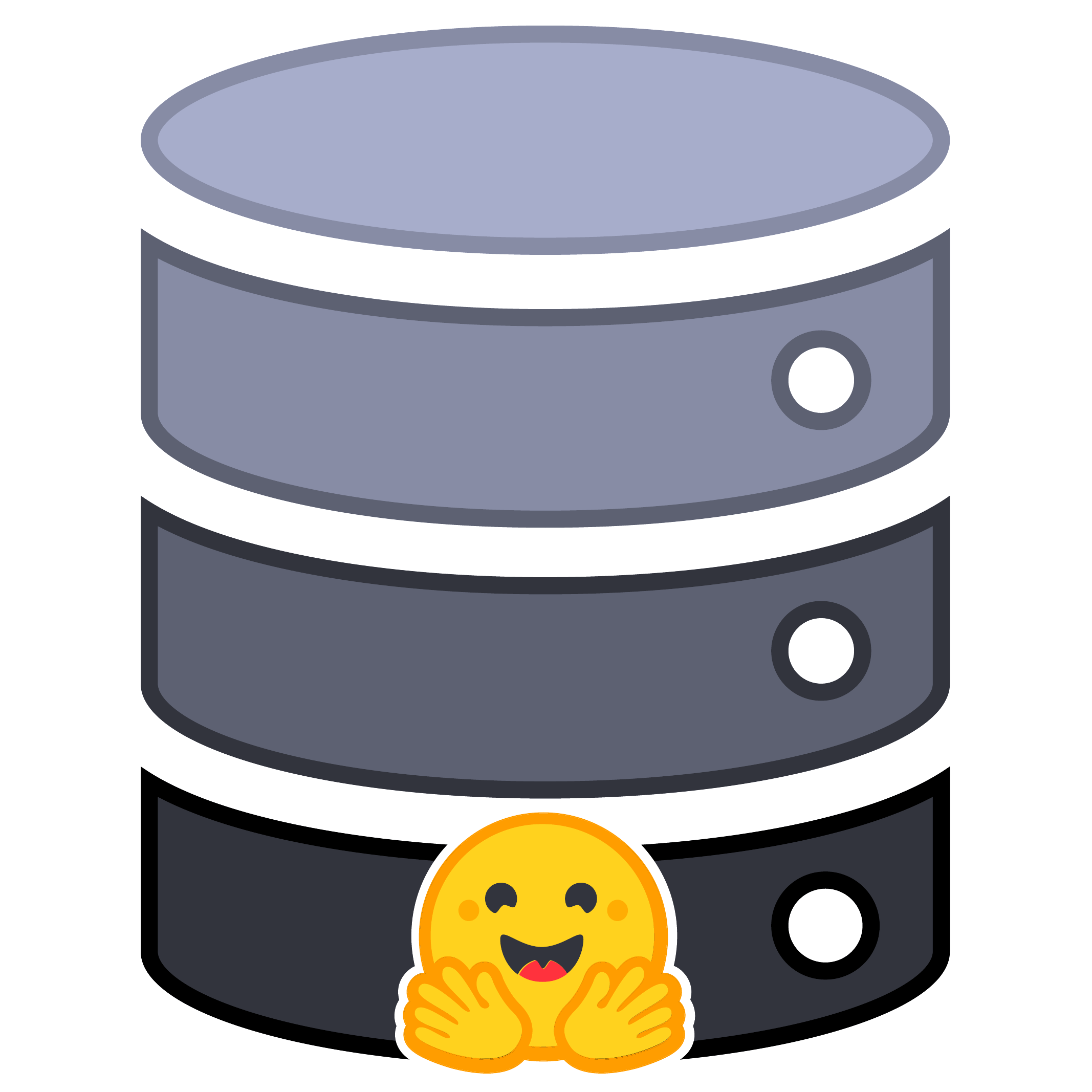}}\xspace}
\newcommand{\github}{\raisebox{-1.5pt}{\includegraphics[height=1.05em]{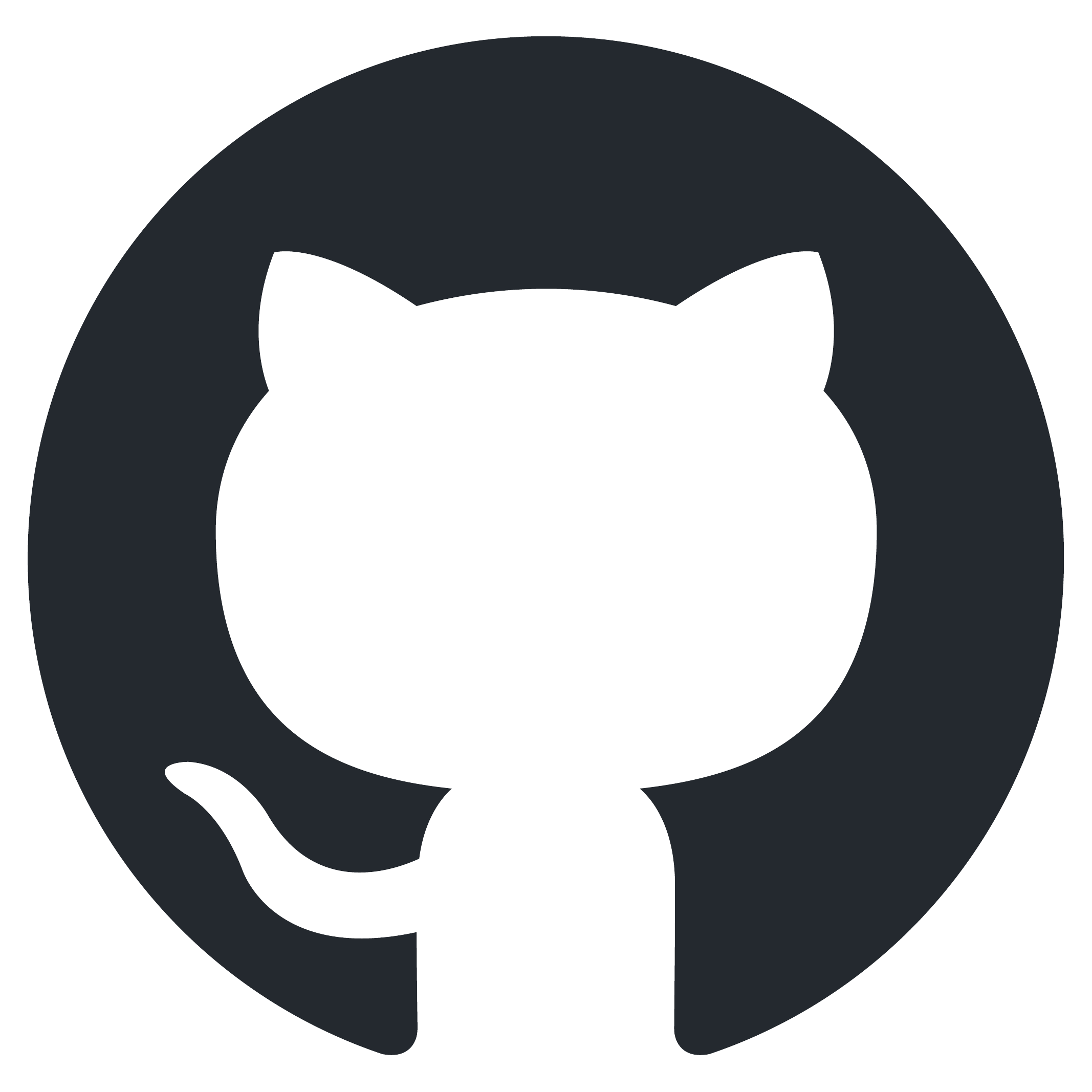}}\xspace}
\newcommand{\webicon}{\raisebox{-1.5pt}{\includegraphics[height=1.05em]{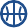}}\xspace}

\definecolor{newyellow}{HTML}{FFD94D}
\definecolor{newgrey}{HTML}{7F7F7F}
\definecolor{newpink}{HTML}{FBCDF4}
\definecolor{realworldoft}{HTML}{029533}
\definecolor{realworldsf}{HTML}{8CC46A}
\definecolor{realworldour}{HTML}{FFC715}
\newcommand{\method}{\textit{RoboMemArena}}

\newcommand{\cmark}{\textcolor[rgb]{0,0.6,0}{$\checkmark$}}
\newcommand{\xmark}{\textcolor{red}{$\times$}}

\tcbuselibrary{theorems}
\newtcbtheorem{finding}{Finding}{
  colback=openhelixred!5,
  colframe=openhelixred!60!black,
  fonttitle=\bfseries,
}{find}

\title{RoboMemArena: A Comprehensive and Challenging Robotic Memory Benchmark}
\author[1,*]{Huashuo Lei}
\author[1,*,\dagger]{Wenxuan Song}
\author[1]{Huarui Zhang}
\author[5]{Jieyuan Pei}
\author[1]{Jiayi Chen}
\author[1]{Haodong Yan}
\author[2,3]{Han Zhao}
\author[2,3]{Pengxiang Ding}
\author[6]{Zhipeng Zhang}
\author[4]{Lida Huang}
\author[3]{Donglin Wang}
\author[4]{Yan Wang}
\author[1,\ddagger]{Haoang Li}

\affiliation[1]{The Hong Kong University of Science and Technology (Guangzhou)}
\affiliation[2]{Zhejiang University}
\affiliation[3]{Westlake University}
\affiliation[4]{Tsinghua University}
\affiliation[5]{Zhejiang University of Technology}
\affiliation[6]{Shanghai Jiao Tong University}

\contribution[*]{Equal Contribution}
\contribution[\dagger]{Project Lead}
\contribution[\ddagger]{Corresponding Author}

\metadata[Code]{\github~\href{https://github.com/OpenHelix-Team/RoboMemArena}{\texttt{RoboMemArena}}}
\metadata[Dataset]{\hfdataset~\href{https://huggingface.co/datasets/RoboMemArenaBenchmark/RoboMemArena}{\texttt{RoboMemArenaBenchmark/RoboMemArena}}}
\metadata[Model Weights]{\huggingface~\href{https://huggingface.co/huashuolei/PrediMem}{\texttt{huashuolei/PrediMem}}}
\metadata[Project Page \& Leaderboard]{\webicon~\href{https://robomemarena.github.io}{\texttt{github.io/RoboMemArena}}}

\abstract{
Memory is a critical component of robotic intelligence, as robots must rely on past observations and actions to accomplish long-horizon tasks in partially observable environments.
However, existing robotic memory benchmarks still lack multimodal annotations for memory formation, provide limited task coverage and structural complexity, and remain restricted to simulation without real-world evaluation.
We address this gap with \textit{RoboMemArena}, a large-scale benchmark of 26 tasks, with average trajectory lengths exceeding 1,000 steps per task and 68.9\% of subtasks being memory-dependent.
The generation pipeline leverages a vision-language model (VLM) to design and compose subtasks, generates full trajectories through atomic functions, and provides memory-related annotations, including subtask instructions and native keyframe annotations, while paired real-world memory tasks support physical evaluation.
We further design \textit{PrediMem}, a dual-system VLA in which a high-level VLM planner manages a memory bank with recent and keyframe buffers and uses a predictive coding head to improve sensitivity to task dynamics.
Extensive experiments on RoboMemArena show that PrediMem outperforms all baselines and provides insights into memory management, model architecture, and scaling laws for complex memory systems.
}

\begin{document}

\maketitle

\afterpage{%
\begin{figure*}[t]
    \centering
    \includegraphics[width=\linewidth]{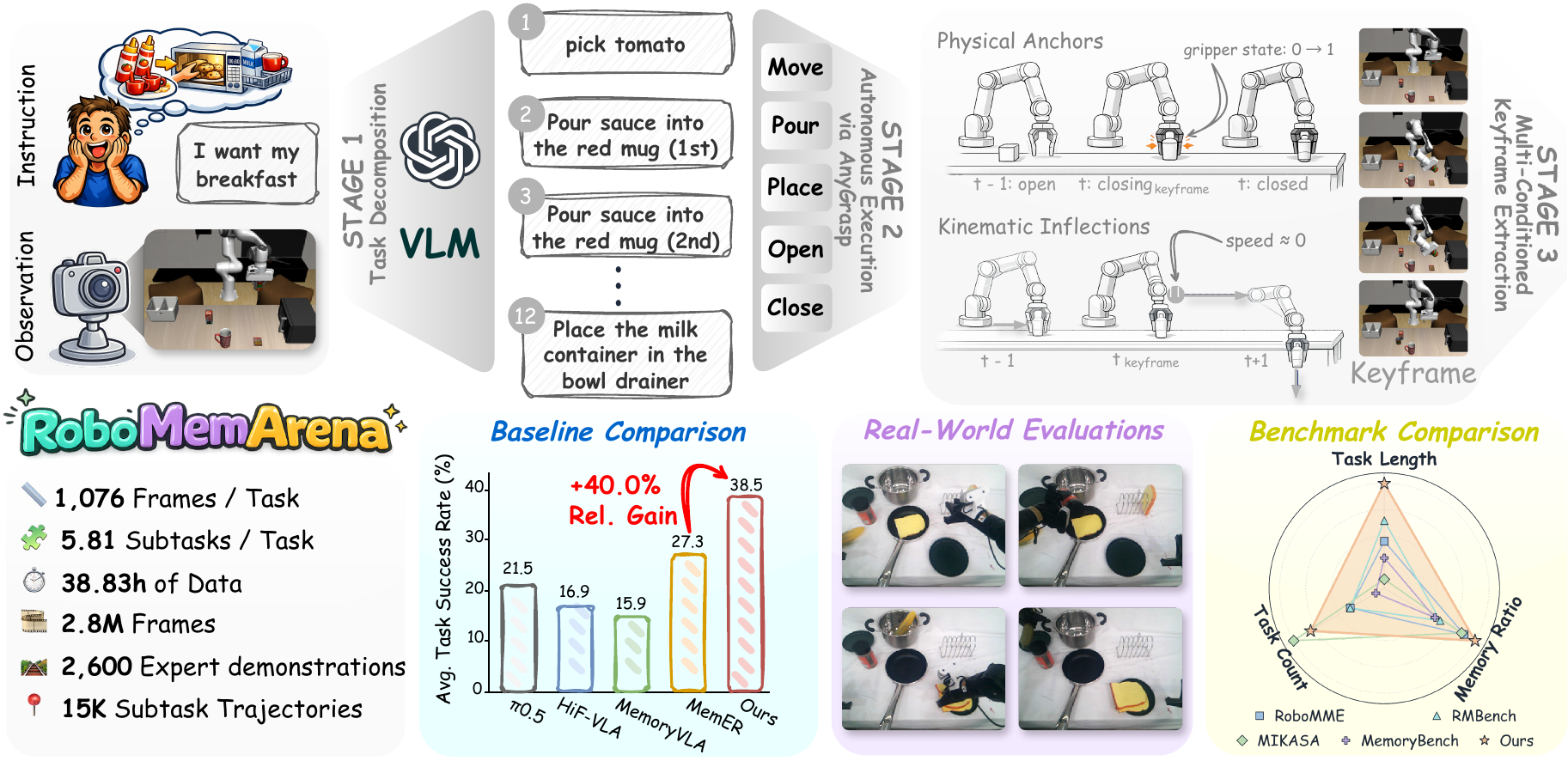}
    \vspace{-15pt}
    \caption{\textbf{Overview of RoboMemArena.}
    Our scalable annotation pipeline converts natural-language instructions into keyframe-annotated trajectories via VLM-based task decomposition, autonomous execution, closed-loop verification, and targeted human refinement of unsuitable annotations.}
    \vspace{-15pt}
    \label{fig:pipeline}
\end{figure*}
}

%

\section{Introduction}
\label{sec:intro}

Memory is a critical component of robotic intelligence, as it determines whether a robot can accomplish long-horizon and complex tasks in partially observable environments. With the advancement of robot foundation policies~\citep{kim2024openvla,intelligence2025pi05visionlanguageactionmodelopenworld,intelligence2025pi06vlalearnsexperience}, recent research~\citep{shi2025memoryvla, sridhar2025memer, lin2025hif, torne2026mem, fang2025sam2act} has begun to endow these foundation models with effective memory mechanisms, enabling them to handle longer-horizon and more complex tasks. 

This trend drives the development of corresponding benchmarks~\citep{fang2025sam2act, cherepanov2025mikasa, rmbench2026}.
However, existing robotic memory benchmarks suffer from several limitations. (1) Their datasets lack the multimodal annotations necessary for memory formation. Recent works~\citep{torne2026mem, intelligence2026pi} have highlighted the inherently multimodal nature of robotic memory. Similar to human memory, comprehensive memory representations may include multiple modalities, such as visual information (\textit{e.g.}, keyframe images) and language (\textit{e.g.}, subtask instructions). Existing benchmarks, however, do not provide such annotations. (2) Their task coverage remains limited: they primarily focus on short-term memory, exhibit relatively low structural complexity, offer limited task diversity, and, in many cases, include tasks that do not genuinely require memory. (3) These benchmarks are restricted to simulation and lack corresponding real-world robotic evaluations. As a result, there remains a significant gap between memory effectiveness in simulated planning and execution in the physical world.


We address this gap with our \textit{RoboMemArena}, a large-scale benchmark built from the ground up for evaluating embodied memory.
In \textit{RoboMemArena}, we design and compose multiple subtasks using a vision-language model (VLM), generate the full trajectory through atomic functions, and subsequently provide memory-related annotations (\textit{i.e.}, subtask instructions and keyframe annotations). This automated pipeline is well-suited for large-scale data generation. 
The simulated benchmark contains 26 tasks across 4 memory-dependent categories (transferring, occlusion, counting, sequential execution), with an average trajectory length of 1,076 steps per task and 68.9\% history-dependent subtasks, which is the highest ratio among existing robotic benchmarks.
As the complementary to the above simulated benchmarks focusing on scalability and reproducibility, we provide real-world benchmarks for physical evaluation.
Specifically, we design 5 challenging real-world memory tasks, with the most complex demonstrations lasting over three minutes.

Furthermore, we design \textbf{PrediMem}, a dual-system VLA that pairs a high-level VLM planner to harness hierarchical memory with a low-level VLA actor.
The VLM manages a memory bank, including a recent buffer and a keyframe buffer.
To enhance the sensitivity to the choice of keyframes, it is combined with a predictive coding head to better understand world dynamics of events and the progression of tasks.
Finally, we conduct extensive experiments of PrediMem on \textit{RoboMemArena} and provide several insights into the memory management, model architecture, and scaling laws of a complex memory system.

In summary, our contributions are:
\begin{itemize}[itemsep=2pt, topsep=2pt, leftmargin=1.5em]
    \item \textbf{Benchmark.} We introduce \textit{RoboMemArena}, a comprehensive and challenging benchmark suitable for validating robotic memory.
    It is equipped with multimodal memory-related annotations, long-horizon and diverse tasks, while supporting real-world tasks. 
    \item \textbf{Model.} We propose \textbf{PrediMem}, a dual-system memory VLA baseline with predictive decoding.
    \item \textbf{Experiments.} We evaluate representative baselines and variants of PrediMem on \textit{RoboMemArena}, showing insights into memory management, model architecture, and scaling laws for memory-augmented robotic manipulation.
\end{itemize}

\section{Related Work}
\label{sec:related}

\subsection{Robotic Memory Benchmarks}

Existing robotic manipulation benchmarks~\citep{xiang2020sapien,mu2025robotwin,nasiriany2024robocasa,tao2025maniskill3gpuparallelizedrobotics,li2024behavior1k,li2024evaluatingrealworldrobotmanipulation,lu2024garmentlab,wang2025dexgarmentlab} cover broad objects, scenes, and skills, but many tasks remain locally observable and therefore do not isolate memory as the central bottleneck. Recent memory-oriented benchmarks~\citep{cherepanov2025mikasa, fang2025sam2act, rmbench2026, dai2026robomme} move closer to this goal, yet three gaps remain. First, they lack rich multimodal memory annotations that can directly supervise dual-system planners. Second, existing memory benchmarks are often limited in task scale and diversity. MemoryBench~\citep{fang2025sam2act} and MIKASA~\citep{cherepanov2025mikasa} are memory-focused, but both remain short-horizon and mainly evaluation-oriented. RMBench~\citep{rmbench2026} broadens memory-complexity settings, but the task coverage is relatively small. Third, most benchmarks are not paired with aligned real-world memory evaluations. RoboMME~\citep{dai2026robomme} standardizes multiple memory dimensions, but its annotations mainly focus on subtask-boundary keyframes and stage-level signals rather than richer multimodal memory supervision. By contrast, \textit{RoboMemArena} addresses these gaps jointly with native multimodal supervision, scalable memory-dependent manipulation tasks, and paired real-world memory evaluations.

\subsection{VLA Models with Memory}

Large-scale VLA pretraining has produced strong language-conditioned manipulation backbones~\citep{kim2024openvla,intelligence2025pi05visionlanguageactionmodelopenworld,intelligence2025pi06vlalearnsexperience,team2024octo,zheng2025xvlasoftpromptedtransformerscalable,chi2025diffusion,zhao2023learningfinegrainedbimanualmanipulation,wen2025dexvla,bai2026hex,cui2025openhelix,song2025rationalvla}, with recent extensions adding multi-frame context and future-aware action modeling~\citep{li2024cogact,li2025cronusvla,lin2025hif,jang2025contextvla,sun2026vla,zhang2026dreamvla,hu2026arvla,song2026reconvla,li2025spatial,zhao2026frappe,song2025accelerating,qiu2026efficient}. We refer to VLAs that predict the next action from the current observation without event memory as \emph{reactive policies}. These policies become brittle when task-relevant information lies in the past. Memory-augmented VLAs~\citep{sridhar2025memer,robocerebra2025,shi2025memoryvla,hu2025adaptiveworkingmemory,koo2025hamlet,li2025mapvla,li2026dualmemoryvla,torne2026mem,sun2026tempofit,wang2026tacmamba,fang2025sam2act,lei2025robomemorybraininspiredmultimemoryagentic,bi2025motus,memctrl2026,star2026,last02026,mempo2026,improving2026} address this limitation through visual retrieval, history reasoning, working memory, temporal caches, or multimodal memory compression. 
For keyframe selection, prior work uses gripper or velocity heuristics, progress-aware embeddings, or retrieved visual keyframes~\citep{james2022coarse,keyframechaining2026,sridhar2025memer}. PrediMem differs by using predictive coding to reshape the shared VLM hidden space, allowing keyframes to be selected through the standard language model (LM) head without extra retrieval modules at inference.

\section{RoboMemArena}
\label{sec:robomemarena}


In this section, we introduce \textit{RoboMemArena}, a complex and challenging robotic memory benchmark,
in four parts:
(1) We present a task suite with four memory-demand categories (\textit{i.e.}, transferring, occlusion, counting, and sequential execution) as well as paired real-world tasks (\Cref{sec:task-setting}).
(2) We propose a data generation pipeline that combines VLM-based task decomposition, autonomous execution, and multi-conditioned keyframe extraction (\Cref{sec:data-pipeline}).
(3) We compare our \textit{RoboMemArena} with existing robotic benchmarks (\Cref{sec:data-analysis}).
(4) We introduce an evaluation protocol that measures both full-task success and stage-level progress (\Cref{sec:eval-protocol}).

\subsection{Task Setting}
\label{sec:task-setting}


\noindent
\textbf{Simulation.}
\textit{RoboMemArena} is designed to evaluate the complementary regime, where the next action depends on task-relevant information that is no longer visible.
The 26 tasks cover four representative failure modes of reactive policies:
\textbf{(1) Multi-Object Transferring.}
The agent relocates multiple objects between visually identical containers and must remember the source--target mapping and which transfers have already been completed.
\textbf{(2) Multi-Object Occlusion.}
The agent places objects into drawers or cabinets that later become visually closed, so it must remember what was placed, where it was placed, and the prior state of each container.
This category is the largest in our benchmark (11~tasks), reflecting how often occlusion causes reactive-policy failures in household settings.
\textbf{(3) Multi-Object Counting.}
The agent must perform an action a specified number of times (\emph{e.g.}, pour exactly twice), even when the scene before consecutive repetitions looks nearly identical.
\textbf{(4) Multi-Object Sequence.}
The correct downstream action depends on an earlier subtask outcome, such as placing a new object into the same container used in a previous step.
The challenge is not only the hidden state, but also resolving references that span multiple operations.
We provide one representative task from each of the four categories in \Cref{fig:task-examples}.
Detailed task-by-task descriptions are summarized in Appendix \Cref{tab:appendix-benchmark-tasks}.

\begin{figure}[!t]
    \centering
    \includegraphics[width=0.99\linewidth]{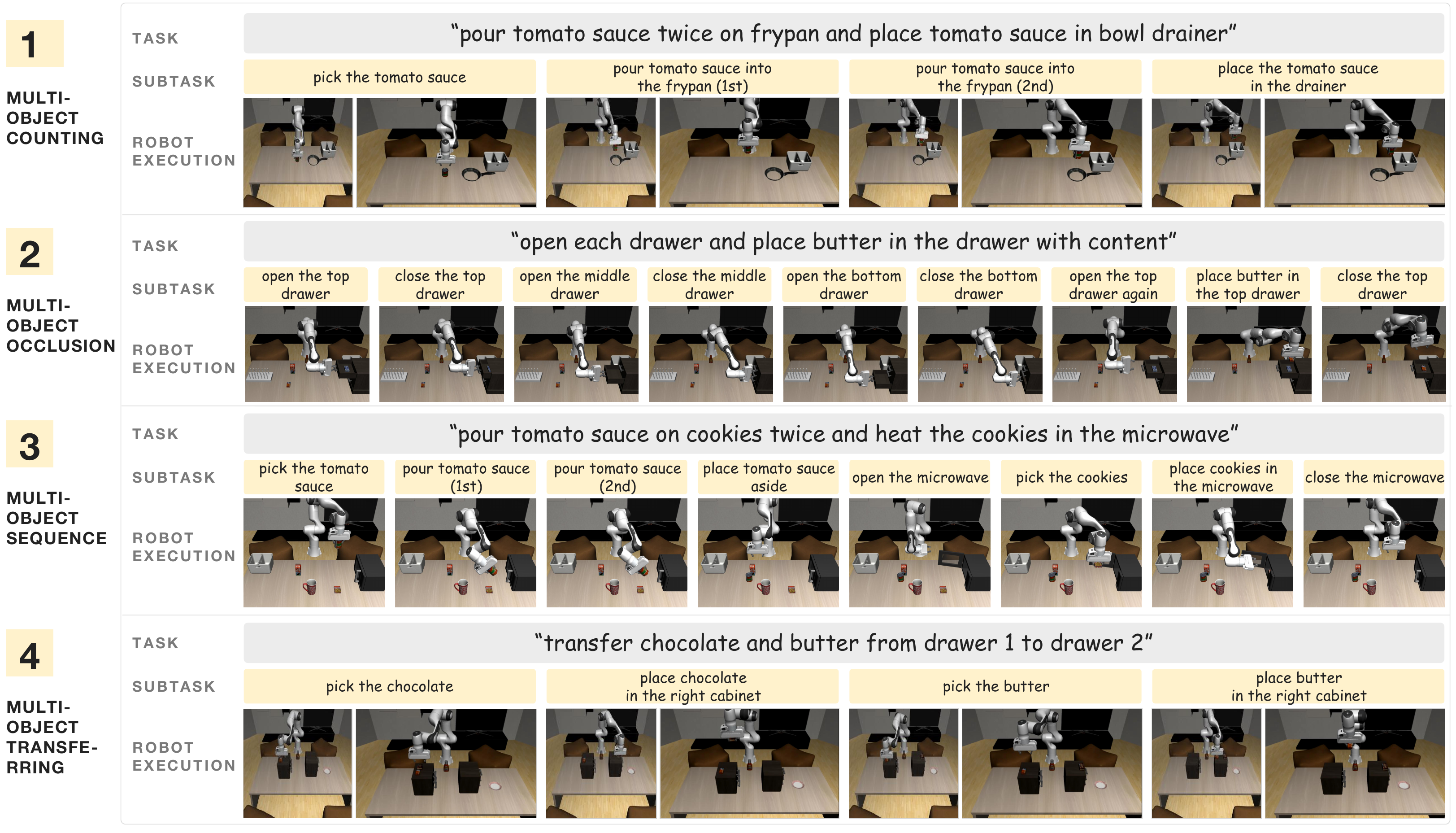}
    \vspace{-10pt}
    \caption{\textbf{Visualization of 4 task categories of RoboMemArena.}
    Each row shows the task instruction, subtask decomposition, and execution rollout for Multi-Object Counting, Occlusion, Sequence, and Transferring.}
    \vspace{-15pt}
    \label{fig:task-examples}
\end{figure}

\noindent
\textbf{Real-world Tasks.}
Beyond simulation, \textit{RoboMemArena} is paired with five real-world memory tasks on a dual-arm platform: Pour Bottle $\times$2, Brush Plates with Swap, Transfer Objects, Shell Game, and Imitate Human to Make Breakfast (IHMB). Together, they cover counting, occlusion, sequential execution, hidden-target tracking, and memory conditioned on human demonstration. 
All tasks are collected and evaluated on the AgileX Cobot Mobile Aloha Platform. 
We use them as a physical validation set for the benchmark design. 
Detailed task descriptions and representative snapshots are provided in Appendix \Cref{tab:appendix-realworld-tasks} and \Cref{fig:real-world-demo}.

\subsection{Automated Data Generation Pipeline}
\label{sec:data-pipeline}


\textit{RoboMemArena} resolves the usual trade-off between scalable automatic collection and fine-grained temporal annotation through three stages (\Cref{fig:pipeline}).

\textbf{Stage 1. VLM-Driven Task Decomposition.}
Given a high-level instruction and the current RGB observation, a VLM proposes an ordered sequence of executable subtasks as scalable initial annotations in simulation.
We then manually refine the subset of decompositions that are unsuitable or inconsistent before downstream execution.
The prompt is designed to expose memory demands such as occlusion, counting, and order-dependent execution.

\textbf{Stage 2. AnyGrasp-Based Autonomous Generation.}
Each subtask is executed autonomously using AnyGrasp~\citep{fang2023anygrasp}, a 6-DoF grasp-pose estimator operating on point-cloud input.
Estimated poses are dispatched to predefined primitives to generate action trajectories.
Moreover, we add a post-condition checker that retries failed subtasks with updated grasp poses.
This closed-loop execution keeps collection automatic while maintaining high success rates.

\textbf{Stage 3. Multi-Conditioned Keyframe Extraction.}
Fixed-frequency sampling either misses state transitions or stores redundant static frames.
Let a continuous trajectory be denoted as $\tau = \{ (s_t, a_t) \}_{t=1}^{T}$, where $s_t$ is the state and $a_t$ is the action at timestep $t$.
We extract the keyframe set $\mathcal{K}$ by taking the union of frames satisfying either of the following two physically grounded conditions:
\begin{equation}
\label{eq:keyframe-union}
\mathcal{K} = \mathcal{K}_{\mathrm{phys}} \cup \mathcal{K}_{\mathrm{kin}}.
\end{equation}

\begin{enumerate}
    \item \textbf{Physical interaction anchors.}
    Gripper-state transitions mark grasp closure and release.
    Let $g_t \in \{0, 1\}$ denote the gripper state (1~=~closed, 0~=~open). The anchor set is:
    \begin{equation}
    \label{eq:k-phys}
    \mathcal{K}_{\mathrm{phys}} = \bigl\{ t \in [1, T] \mid g_t \neq g_{t-1} \bigr\}.
    \end{equation}

    \item \textbf{Kinematic inflections.}
    End-effector velocity minima and abrupt direction changes mark transitions between motion phases.
    Let $\mathbf{v}_t \in \mathbb{R}^3$ be the end-effector linear velocity. We identify a kinematic inflection at timestep $t$ if the velocity magnitude drops below a threshold $\epsilon$ or the cosine similarity between consecutive velocity vectors falls below $\cos(\theta)$:
    \begin{equation}
    \label{eq:k-kin}
    \mathcal{K}_{\mathrm{kin}} = \left\{ t \in [1, T] \ \middle| \ \|\mathbf{v}_t\| < \epsilon \ \lor \ \frac{\mathbf{v}_t \cdot \mathbf{v}_{t-1}}{\|\mathbf{v}_t\| \, \|\mathbf{v}_{t-1}\|} < \cos(\theta) \right\}.
    \end{equation}
\end{enumerate}

Together, these conditions select information-bottleneck frames that reconstruct task progress while avoiding dense video storage.
The annotations provide temporal supervision for VLMs while keeping the memory representation compact and event-focused.

\begin{table}[!t]
    \caption{\textbf{Comparison with Popular Benchmarks used in the Robot Learning Literature.} \textit{RoboMemArena} features long-horizon memory tasks, multimodal memory-related annotations, scalable generation pipelines, and paired real-world evaluations.
    }
    \label{tab:benchmark-comparison}
    \centering
    \small
    \setlength{\tabcolsep}{4pt}
    \resizebox{\linewidth}{!}{
    \begin{tabular}{lcccccccc}
        \toprule
        \textbf{Benchmarks} & \textbf{Long Horizon} & \textbf{Auto Instr. Gen.} & \textbf{Atomic Subgoals} & \textbf{Scalable Gen.} & \textbf{Autonomous Grasp} & \textbf{State Oracle} & \textbf{Native Keyframes} & \textbf{Real-World} \\
        \midrule
        RLBench~\citep{james2020rlbench}       & \xmark & \xmark & \xmark & \cmark & \xmark & \xmark & \xmark & \xmark \\
        RoboCerebra~\citep{robocerebra2025}    & \cmark & \cmark & \cmark & \xmark & \xmark & \cmark & \xmark & \xmark \\
        ARNOLD~\citep{gong2023arnold}          & \xmark & \xmark & \xmark & \cmark & \xmark & \xmark & \xmark & \xmark \\
        ALFRED~\citep{shridhar2020alfred}      & \xmark & \xmark & \cmark & \cmark & \xmark & \cmark & \xmark & \xmark \\
        CALVIN~\citep{mees2022calvin}          & \xmark & \xmark & \xmark & \xmark & \xmark & \xmark & \xmark & \xmark \\
        RoboCasa~\citep{nasiriany2024robocasa} & \xmark & \cmark & \cmark & \cmark & \xmark & \cmark & \xmark & \cmark \\
        LIBERO-Long~\citep{liu2023libero}      & \xmark & \xmark & \xmark & \xmark & \xmark & \xmark & \xmark & \xmark \\
        VLABench~\citep{zhang2025vlabench}     & \xmark & \cmark & \cmark & \cmark & \xmark & \cmark & \xmark & \xmark \\
        RoboTwin~\citep{mu2025robotwin}        & \xmark & \cmark & \xmark & \cmark & \xmark & \xmark & \xmark & \cmark \\
        RMBench~\citep{rmbench2026}            & \xmark & \cmark & \cmark & \cmark & \xmark & \cmark & \cmark & \cmark \\
        RoboMME~\citep{dai2026robomme}         & \xmark & \xmark & \cmark & \cmark & \xmark & \cmark & \cmark & \cmark \\
        BEHAVIOR-1K~\citep{li2024behavior1k}   & \cmark & \xmark & \cmark & \xmark & \xmark & \cmark & \xmark & \cmark \\
        MIKASA~\citep{cherepanov2025mikasa}    & \xmark & \xmark & \cmark & \xmark & \xmark & \cmark & \xmark & \cmark \\
        MemoryBench~\citep{fang2025sam2act}    & \xmark & \xmark & \cmark & \xmark & \xmark & \cmark & \xmark & \cmark \\
        \textbf{RoboMemArena~(Ours)}          & \textbf{\cmark} & \textbf{\cmark} & \textbf{\cmark} & \textbf{\cmark} & \textbf{\cmark} & \textbf{\cmark} & \textbf{\cmark} & \textbf{\cmark} \\

        \bottomrule
    \end{tabular}
    }

    \vspace{1mm}
    \raggedright \scriptsize
    \textit{Note:}
    \textbf{Long Horizon}: Tasks are categorized as long-horizon when the average trajectory length is greater than 1000 steps.
    \textbf{Auto Instr. Gen.}: Automated generation or augmentation of task instructions using language models, vision-language models, or multimodal models.
    \textbf{Atomic Subgoals}: Tasks provide explicit step-level subgoals, subtask annotations, or symbolic atomic goal predicates beyond a single final goal.
    \textbf{Scalable Gen.}: Automated batch generation of executable trajectories, demonstrations, or dense action annotations in simulation, rather than manual trajectory collection.
    \textbf{Autonomous Grasp}: Autonomous grasping through point-cloud-based grasp pose estimation, enabling executable grasp actions without manual grasp annotations.
    \textbf{State Oracle}: Programmatic signals for judging subtask or stage completion from simulator execution traces, object states, or symbolic predicates.
    \textbf{Native Keyframes}: Explicit temporal keyframe dataset construction for hierarchical supervision or memory-oriented evaluation.
    \textbf{Real-World Eval.}: The benchmark paper or project includes explicit real-robot evaluation associated with the benchmark setup.
    \vspace{-0.4cm}
\end{table}

\subsection{Data Analysis}
\label{sec:data-analysis}


To highlight the unique features of \textit{RoboMemArena}, we perform qualitative comparisons against popular robotic benchmarks and quantitative comparisons against existing robotic memory benchmarks.

\textbf{Benchmark Comparison.}
We provide a thorough comparison between \textit{RoboMemArena} and 14 established benchmarks across 8 feature dimensions in \Cref{tab:benchmark-comparison}.  
\textit{RoboMemArena} is the only entry that satisfies all eight criteria.
Taken together, these comparisons highlight three benchmark-level strengths of \textit{RoboMemArena}: richer multimodal memory supervision through native keyframes, broader scale and diversity through automated trajectory generation, and paired real-world evaluation for physical validation.

\begin{figure}[!t]
    \centering
    \includegraphics[width=\linewidth]{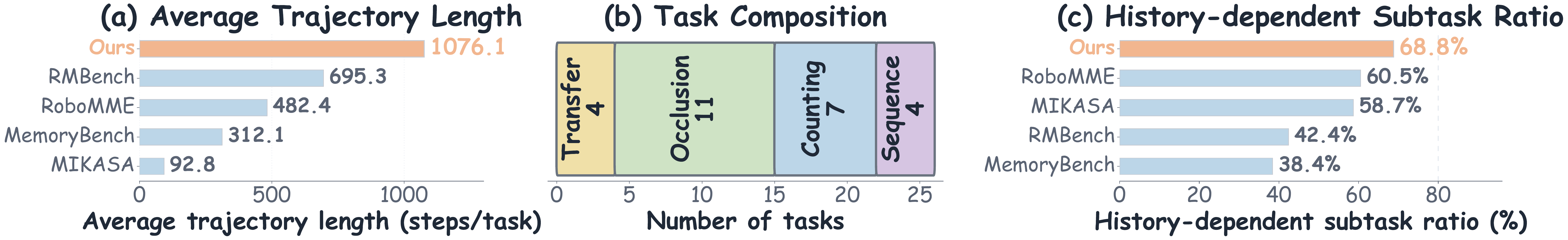}
    \vspace{-20pt}
    \caption{\textbf{Summary statistics of RoboMemArena.}
    Panels (a)--(c) respectively illustrate the average trajectory length, task
    composition, and history-dependent subtask ratio, highlighting
    the long-horizon nature of \textit{RoboMemArena} and the prevalence of
    memory-conditioned subtasks relative to prior benchmarks.}
    \vspace{-15pt}
    \label{fig:stats}
\end{figure}

\textbf{Memory-Dependent Subtask Ratio.}
In total, \textit{RoboMemArena} defines 151~distinct subtasks across its 26~tasks.
We consider a subtask memory-dependent if its correct execution cannot be inferred from the current observation alone and requires information from earlier subtasks or observations.
For the $i$-th task with $n_i$ subtasks and $m_i$ memory-dependent subtasks, its task-level memory ratio is $r_i=m_i/n_i$.
Across all tasks in \textit{RoboMemArena}, 104 of 151~subtasks are memory-dependent, giving a \textbf{68.9\%} history-dependent subtask ratio. 
\Cref{fig:stats}(c) shows \textit{RoboMemArena} also has the highest history-dependent subtask ratio among all robotic memory benchmarks.
The calculating protocol is detailed in Appendix \Cref{sec:appendix-memory-ratio}.


\textbf{Scale and Diversity.}
For each of the 26 tasks, we collect 100 successful demonstrations, yielding 2,600 long-horizon visual trajectories. These produce 15,100 keyframe-aligned short segments for hierarchical supervision.
In terms of average trajectory length, our \textit{RoboMemArena} is longer than existing robotic memory benchmarks, achieving 1,076 steps per task, as shown in \Cref{fig:stats}(a). \Cref{fig:stats}(b) shows the task composition of \textit{RoboMemArena}, which includes 4 transferring tasks, 11 occlusion tasks, 7 counting tasks, and 4 sequence tasks.

\subsection{Evaluation Protocol}
\label{sec:eval-protocol}


Binary success alone is insufficiently informative for long-horizon memory tasks. Therefore, we report both full-task success and partial progress.

\textbf{Task Success Rate (TSR).}
To determine whether a task is successful, we verify it through multiple stage-level predicates rather than only checking the final outcome.
For the $i$-th task, we define $K_i$ stage-level verification predicates $\psi(s_i^{(k)})$ for $k = 1, \dots, K_i$, where $s_i^{(k)}$ denotes the execution state at the $k$-th verification stage.
These predicates encode state conditions such as object location, containment, visibility, and stage completion.
Each predicate returns $\texttt{True}$ if the corresponding condition holds.
A task is deemed successful only when \emph{all} predicates are satisfied:
\begin{equation}
\label{eq:tsr}
\mathrm{TSR} = \frac{1}{N} \sum_{i=1}^{N} \prod_{k=1}^{K_i} \mathbf{1}\!\left[\psi\!\left(s_i^{(k)}\right)\right],
\end{equation}
where $N$ is the total number of evaluated tasks, and $\mathbf{1}[\cdot]$ denotes the indicator function, which equals 1 if the predicate is satisfied and 0 otherwise.

\textbf{Cumulative Success Rate (CSR).}
Rather than requiring all-or-nothing success, CSR measures the \emph{fraction} of verification stages that each task completes, thereby quantifying task progress:
\begin{equation}
\label{eq:csr}
\mathrm{CSR} = \frac{1}{N} \sum_{i=1}^{N} \frac{1}{K_i} \sum_{k=1}^{K_i} \mathbf{1}\!\left[\psi\!\left(s_i^{(k)}\right)\right].
\end{equation}
CSR distinguishes partial completion from complete failure.
Appendix \Cref{fig:verification} shows that the number of verification stages per task ranges from 3 to 9, and the majority of tasks exceed 5.
This distribution gives CSR enough resolution to compare memory degradation across temporal horizons.

%

\section{PrediMem: Building Hierarchical Memory with Predictive Coding}
\label{sec:predimem}



\begin{figure}[t]
    \centering
    \includegraphics[width=0.99\linewidth]{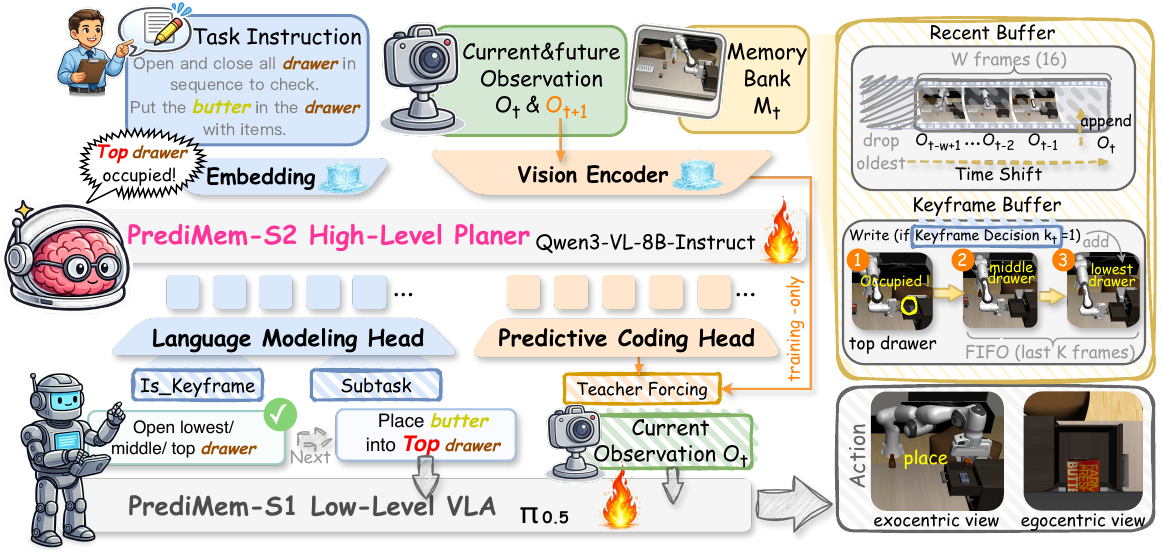}
    \vspace{-10pt}
    \caption{\textbf{The PrediMem pipeline.}
    The pipeline comprises two asynchronously coupled components: S1, a low-level action policy that executes the current subtask, and S2, a high-level planner that predicts keyframes and dispatches the next subtask. The predictive coding head path is training-only.}
    \vspace{-0.5cm}
    \label{fig:framework}
\end{figure}

We introduce \textbf{PrediMem}, a hierarchical \textbf{Mem}ory framework with \textbf{Predi}ctive coding for embodied memory. It consists of a high-level planner (System 2, denoted by S2), a low-level execution policy (System 1, denoted by S1), a keyframe-grounded memory bank, and an auxiliary predictive coding head.
As shown in \Cref{fig:framework}, the memory bank $\mathcal{M}_t$ combines a long-term keyframe buffer $\mathcal{M}_t^{\mathrm{key}}$ with a recent sliding window with fixed horizon $\mathcal{M}_t^{\mathrm{rec}}$, \textit{i.e.}, $\mathcal{M}_{t} = \mathcal{M}_t^{\mathrm{key}} \cup \mathcal{M}_t^{\mathrm{rec}}$.
S2 takes the current observation together with the memory bank to predict the current subtask and decide whether the current frame should be stored as a keyframe. 
Accepted keyframes are written back into the keyframe buffer, allowing the system to preserve decision-critical events beyond the recent observation window. 
Meanwhile, S1 predicts the freshest subtask-conditioned action chunk.


\textbf{Predictive Coding.}
\label{sec:lfp}
The key question is \emph{when} to write frames to memory: over-storing wastes capacity, while missed transitions cause downstream errors.
To enhance the model's sensitivity to keyframes and ability to capture future dynamics, we introduce predictive coding.
Its objective is to predict the representation of the subsequent frame from the visual features of the current frame $o_t$, thereby enabling the model to better capture abrupt state transitions at keyframes.
To this end, we incorporate an additional predictive coding head $f_{\mathrm{Pre}}$ that predicts the visual feature of the subsequent frame $\hat{Z}_{t+1} = f_{\mathrm{Pre}}(h_t)$, with supervision $Z_{t+1}$ provided by the visual encoder of the VLM, \textit{i.e.}, a frozen ViT.
Following Cambrian-S~\citep{yang2025cambrian}, the predictive loss $\mathcal{L}_{\mathrm{Pre}}$ is formulated as the sum of a latent Mean Squared Error term and a cosine-distance term against the stop-gradient teacher next-frame latent features:
\begin{equation}
\label{eq:lfp-loss}
\mathcal{L}_{\mathrm{Pre}}
=
\mathrm{MSE}\!\bigl(\hat{Z}_{t+1}, \mathrm{sg}(Z_{t+1})\bigr)
+
\bigl(1 - \cos\!\bigl(\hat{Z}_{t+1}, \mathrm{sg}(Z_{t+1})\bigr)\bigr).
\end{equation}


%
%

\textbf{Total Training Loss.}
The final objective for S2 combines next-token prediction with the predictive coding loss $\mathcal{L}_\text{S2} = \mathcal{L}_{\mathrm{text}} + 0.1\mathcal{L}_{\mathrm{Pre}}$.
Here, $\mathrm{sg}(\cdot)$ denotes the stop-gradient operator, and $\mathcal{L}_{\mathrm{text}}$ denotes the next-token prediction loss for subtask generation and keyframe decisions.
The loss function used for S1 follows the official flow-matching objective introduced in \citep{intelligence2025pi05visionlanguageactionmodelopenworld}.



\textbf{Inference.}
During inference, the predictive coding head is removed, so our PrediMem retains the architecture and cost of a standard dual-system framework while inheriting the improved capabilities for dynamics understanding and keyframe selection.
The dual system executes asynchronous inference, detailed in Appendix \Cref{sec:appendix-protocol,tab:async-runtime}.


%

\section{Experiments}
\label{sec:exp}
\definecolor{qcolor}{HTML}{1F4E8C}  
We evaluate \textit{RoboMemArena} and the \textbf{PrediMem} framework around five questions:
\begin{enumerate}[leftmargin=2em, itemsep=0pt, topsep=2pt, parsep=0pt]
    \item[\textcolor{qcolor}{\textbf{Q1.}}] Does \textit{RoboMemArena} expose a memory gap in existing VLAs, and can PrediMem close it?
    \item[\textcolor{qcolor}{\textbf{Q2.}}] Does the end-to-end trained robot memory system surpass powerful closed-source agents?
    \item[\textcolor{qcolor}{\textbf{Q3.}}] How much do the predictive coding head and the keyframe bank contribute, and how does predictive coding shape the learned memory representations?
    \item[\textcolor{qcolor}{\textbf{Q4.}}] How does the scaling of memory influence model performance?
    \item[\textcolor{qcolor}{\textbf{Q5.}}] How do different baselines perform in the real-world evaluation of \textit{RoboMemArena}?
\end{enumerate}

\subsection{Experimental Setup}
\label{sec:exp-setup}


\textbf{Baselines.}
We compare against $\boldsymbol{\pi_{0.5}}$~\citep{intelligence2025pi05visionlanguageactionmodelopenworld}, which is a reactive VLA that acts only on the current observation.
We also compare with \textbf{HiF-VLA}~\citep{lin2025hif}, which models hindsight, insight, and foresight motion representations, \textbf{MemoryVLA}~\citep{shi2025memoryvla}, which uses token-level working memory, and \textbf{MemER}~\citep{sridhar2025memer}, which follows a dual-system design with keyframe retrieval.

\textbf{Implementation.}
All experiments are conducted on \textit{RoboMemArena}. We report Task Success Rate (TSR) and Cumulative Success Rate (CSR) as defined in \Cref{sec:eval-protocol}.
PrediMem builds on Qwen3-VL-8B-Instruct with the vision tower frozen and the remaining modules fully fine-tuned for 2 epochs on $4{\times}$H100 with learning rate $1{\times}10^{-5}$. The predictive coding head uses latent MSE and cosine losses (weight $0.1$ each). The recent buffer holds 5 frames, and the keyframe buffer is uncapped. The prompt format is given in Appendix \Cref{sec:appendix-json}.

\subsection{Main Results in Simulation \textcolor{qcolor}{(\textbf{Q1}, \textcolor{qcolor}{\textbf{Q2}})}}
\label{sec:main-results}

\begin{table*}[t]
\caption{\textbf{Comparison on RoboMemArena.} Per-category TSR/CSR (\%) and overall averages.
Ground Truth shown in gray as an oracle reference.}
\label{tab:main-results}
\centering
\footnotesize
\setlength{\tabcolsep}{4pt}
\renewcommand{\arraystretch}{0.85}
\resizebox{\linewidth}{!}{%
\begin{tabular}{l|cccccccccc}
\toprule
\multicolumn{1}{c|}{\textbf{Method}} & \multicolumn{2}{c}{\textbf{Transferring}} & \multicolumn{2}{c}{\textbf{Occlusion}} & \multicolumn{2}{c}{\textbf{Counting}} & \multicolumn{2}{c}{\textbf{Sequence}} & \multicolumn{2}{c}{\textbf{Average}} \\
\cmidrule(lr){2-3}\cmidrule(lr){4-5}\cmidrule(lr){6-7}\cmidrule(lr){8-9}\cmidrule(lr){10-11}
 & \textbf{TSR} & \textbf{CSR} & \textbf{TSR} & \textbf{CSR} & \textbf{TSR} & \textbf{CSR} & \textbf{TSR} & \textbf{CSR} & \textbf{TSR} & \textbf{CSR} \\
\midrule
\multicolumn{11}{l}{\textit{(a) Baselines}} \\
$\pi_{0.5}$ & 20.0 & 42.8 & 12.7 & 17.2 & 14.3 & 50.9 & 60.0 & 71.6 & 21.5 & 38.7 \\
HiF-VLA & 17.5 & 38.9 & 12.7 & 27.1 & 8.6 & 45.9 & 42.5 & 70.2 & 16.9 & 39.8 \\
MemoryVLA & 15.0 & 37.2 & 7.3 & 13.1 & 14.3 & 55.1 & 37.5 & 65.2 & 15.0 & 35.3 \\
MemER & 20.0 & 36.1 & 16.4 & 33.2 & 27.1 & 65.1 & 65.0 & 79.1 & 27.3 & 49.1 \\
\midrule
\multicolumn{11}{l}{\textit{(b) Frozen / oracle references}} \\
Qwen3-VL-8B (frozen) & 15.0 & 34.6 & 0.0 & 6.8 & 9.3 & 44.6 & 7.5 & 39.2 & 6.0 & 26.2 \\
GPT-5.4 & 13.8 & 32.9 & 1.8 & 9.2 & 12.9 & 50.7 & 15.0 & 47.3 & 8.7 & 30.5 \\
\textcolor{gray}{Ground Truth} & \textcolor{gray}{32.5} & \textcolor{gray}{54.8} & \textcolor{gray}{33.6} & \textcolor{gray}{49.8} & \textcolor{gray}{51.4} & \textcolor{gray}{75.6} & \textcolor{gray}{85.0} & \textcolor{gray}{92.3} & \textcolor{gray}{46.1} & \textcolor{gray}{64.8} \\
\midrule
\multicolumn{11}{l}{\textit{(c) Component variants of PrediMem}} \\
w/o Predictive Coding Head & 25.0 & 43.7 & 19.5 & 30.2 & 38.6 & 61.8 & 63.8 & 80.7 & 32.3 & 49.0 \\
w/o Keyframe Bank & 17.5 & 33.3 & 6.4 & 22.8 & 20.0 & 61.7 & 45.0 & 66.3 & 17.7 & 41.6 \\
\midrule
\multicolumn{11}{l}{\textit{(d) Backbone variants of PrediMem}} \\
w/ Qwen3-1.7B & 15.0 & 31.6 & 7.3 & 20.8 & 28.6 & 60.2 & 50.0 & 73.9 & 19.9 & 41.4 \\
w/ Qwen3-4B & 20.0 & 42.1 & 18.2 & 34.7 & 38.6 & 64.9 & 65.0 & 84.6 & 31.9 & 51.7 \\
\midrule
\textbf{PrediMem (Ours)} & \textbf{22.5} & \textbf{45.2} & \textbf{27.3} & \textbf{38.4} & \textbf{45.7} & \textbf{69.3} & \textbf{72.5} & \textbf{89.5} & \textbf{38.5} & \textbf{55.2} \\
\bottomrule
\end{tabular}
}

\end{table*}

\Cref{tab:main-results}(a) shows $\pi_{0.5}$ reaches 21.5\% average TSR and 38.7\% average CSR.
Because $\pi_{0.5}$ is a reactive policy without explicit history modeling, it attains relatively high success rates only on certain sequential tasks, where many intermediate steps remain governed by local visual regularities and do not inherently require recalling distant past events.
However, when a task requires remembering an earlier state, it fails almost completely.
For example, in drawer-based tasks, once the first drawer has been opened and the scene returns to a visually similar state, the policy may repeatedly revisit or reopen the same drawer instead of tracking which drawer has already been inspected.
HiF-VLA introduces richer motion representations, which help short-horizon execution, but it does not explicitly store task-level events such as object placements, previous drawer states, or action counts.
MemoryVLA stores history as transformer tokens, but this token-level memory is not explicitly aligned with the sparse physical transitions that determine task progress in our benchmark.
MemER performs better than purely reactive baselines because it retrieves visual keyframes and follows a dual-system design.
However, constrained by its limited perception of task dynamics, the high-level VLM is often unable to select informative keyframes.

In comparison, PrediMem combines an explicit keyframe bank with predictive coding.
The keyframe bank preserves task-relevant events beyond the recent observation window, while predictive coding makes the high-level VLM more sensitive to physical state transitions.
This hierarchical memory input and precise subtask management bring the highest TSR and CSR among all methods.

\textbf{End-to-end Training \textit{v.s.} Closed-sourced VLMs as S2.}
Despite their strong memory capabilities in language and multimodal domains, closed-source agents transfer poorly to robotic memory-intensive tasks.
\Cref{tab:main-results}(b) shows that closed-sourced VLMs fall far below trainable baselines, even the SOTA GPT-5.4~\citep{openai2026gpt54} achieves only 8.7\% TSR.
We attribute this gap to limited memory generalization to unseen robotic scenarios and an insufficient understanding of physical actions in VLMs trained primarily on vision-language tasks. These results suggest that VLMs must be trained on robotic data to serve effectively as memory management modules in robotic systems.

\subsection{Ablation Studies \textcolor{qcolor}{(\textcolor{qcolor}{\textbf{Q3}})}}
\label{sec:ablation}


\textbf{Predictive Coding.}
\Cref{tab:main-results}(c) shows the ablation results of predictive coding and the keyframe bank. 
For transferring tasks, removing the predictive coding head has a limited effect because its relevant state changes are direct. 
In contrast, predictive coding becomes more important for occlusion, counting, and sequence tasks, where the model must detect subtle state transitions such as drawer closure, object disappearance, repeated pouring, or the completion of an ordered step.
By predicting future visual representations during training, predictive coding makes the hidden states more sensitive to these transition points and can recover discriminative cues beyond the sparsely annotated keyframes, mitigating imperfections in both keyframe labeling and SFT supervision.

\begin{wraptable}{r}{0.45\textwidth}
\vspace{-20pt}
\centering
\footnotesize
\setlength{\tabcolsep}{10pt}
\renewcommand{\arraystretch}{1.0}
\caption{\textbf{Ablations of $\mathcal{L}_{\mathrm{Pre}}$ Weights.}}
\resizebox{\linewidth}{!}{%
\begin{tabular}{l|cccc}
\toprule
Weights
& 0.0
& 0.1
& 0.5
& 1
 \\
\midrule
TSR        & 32.3\% & \textbf{38.5\%} & 31.0\%          & 29.8\%  \\
\bottomrule
\end{tabular}
}
\label{tab:loss_weight}
\vspace{-0.5cm}
\end{wraptable}
\textbf{Loss Weights.}
A coefficient is introduced to balance the predictive loss $\mathcal{L}_{\mathrm{Pre}}$ and the instruction-tuning loss $\mathcal{L}_{\mathrm{text}}$. As shown in \Cref{tab:loss_weight}, our ablation study indicates that 0.1 yields the best performance.

\textbf{Keyframe Bank.}
Removing the keyframe bank also causes a broader drop because task-relevant events are no longer preserved once they leave the recent-frame window.
For occlusion and sequence tasks, the correct decision often depends on earlier placements, inspections, or ordering decisions that are no longer visible.
Thus, explicit keyframe memory is necessary for accurate high-level planning.

\textbf{Visualization of Keyframe Representation.}
As qualitative studies, \Cref{fig:analysis-triptych}(c) visualizes keyframe-related hidden representations on Task~1 using t-SNE.
We extract final-layer hidden representations for the same input samples, aggregate token features into one embedding per sample, and project these embeddings to two dimensions for visualization.
Without predictive coding, embeddings from different keyframe classes overlap strongly. With predictive coding, samples from the same class become more compact and different classes are more separated.
This indicates that predictive coding learns more discriminative keyframe representations, which contribute to making more precise subsequent keyframe decisions.
\subsection{Scaling Studies \textcolor{qcolor}{(\textcolor{qcolor}{\textbf{Q4}})}}
\label{sec:memory-capacity}

\begin{figure*}[t]
    \centering
    \includegraphics[width=\linewidth]{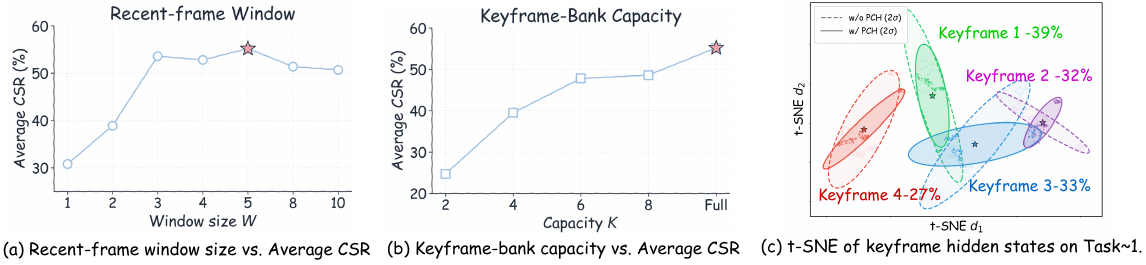}
    \vspace{-20pt}
    \caption{\textbf{Analyses of memory behavior.}
    (a) and (b) shows the sensitivity of average CSR to the recent-buffer size and keyframe-bank capacity. 
    (c) is a t-SNE view showing that predictive coding yields tighter, more discriminative keyframe clusters.}
    \vspace{-15pt}
    \label{fig:analysis-triptych}
\end{figure*}

\textbf{Scaling of Memory Bank.}
\Cref{fig:analysis-triptych}(a,b) summarizes how performance changes as the recent-frame buffer and keyframe buffer are varied.
With only one or two recent frames, S1 lacks enough temporal evidence to detect state transitions and select keyframes reliably.
A 3--5 frame window is sufficient for most short-term changes, while larger windows add redundant visual context, increase VLM latency, and make S1 refreshes less synchronized with the low-level VLA execution loop.
For the keyframe bank, CSR is very low with only two stored keyframes because early observations are quickly evicted in long-horizon tasks.
Increasing the capacity to 4--8 improves performance, and the best result is obtained with the uncapped keyframe bank, which preserves early decision-critical observations such as the first drawer state after later drawers have been inspected.

\textbf{Scaling of S2.}
In \Cref{tab:main-results}(d), we train and evaluate our method using Qwen3 backbones with different parameters. 
Under controlled settings with matched pretraining data and model architecture, we find that scaling up the S2 model consistently improves performance across all tasks. 
This trend suggests that larger models provide stronger reasoning and memory capabilities, which are particularly beneficial for our benchmark.

\subsection{Real-World Experiments \textcolor{qcolor}{(\textcolor{qcolor}{\textbf{Q5}})}}
\label{sec:real-world}

\begin{wraptable}{r}{0.5\textwidth}
\vspace{-2cm}
\centering
\footnotesize
\setlength{\tabcolsep}{3pt}
\renewcommand{\arraystretch}{1.0}
\caption{\textbf{Real-world success rates (\%).}
Each task is evaluated over 10 rollouts on a dual-arm platform.}
\resizebox{\linewidth}{!}{%
\begin{tabular}{l|ccccc|c}
\toprule
\textbf{Method}
& \textbf{Pour}$\times$2
& \textbf{Brush}
& \textbf{Transfer}
& \textbf{Shell}
& \textbf{IHMB}
& \textbf{Avg.} \\
\midrule
$\pi_{0.5}$        & 20 & 10 & 60          & 10 & 0           & 20 \\
MemER          & 30 & 50 & \textbf{80} & 40 & 0           & 40 \\
\midrule
\textbf{PrediMem}  & \textbf{60} & \textbf{60} & \textbf{80} & \textbf{50} & \textbf{10} & \textbf{52} \\
\bottomrule
\end{tabular}
}
\label{tab:real-world}
\vspace{-10pt}
\end{wraptable}

We compare PrediMem against $\pi_{0.5}$ and MemER on the five real-world memory tasks of \textit{RoboMemArena}, introduced in \Cref{sec:task-setting}.
\Cref{tab:real-world} shows $\pi_{0.5}$ achieves only 20\% average success because it selects actions from the current frame alone and cannot reliably use previous counts, hidden target locations, or demonstrated action order.
MemER improves the average success rate from 20\% to 40\%, benefiting from the valuable high-level memory signal provided by retrieved keyframes.
PrediMem further improves the average success rate to 52\%, outperforming MemER on four tasks
Please note that for the 3-minute longest-horizon task, Imitate Human to Make Breakfast (IHMB), only our PrediMem succeeds. This result demonstrates that, in complex real-world scenarios, designing an effective memory mechanism is critical for successful task completion.



%

\section{Conclusion}
\label{sec:conclusion}

We presented \textit{RoboMemArena}, a diverse and challenging robotic memory benchmark that combines native keyframe-centered multimodal supervision, scalable long-horizon trajectory generation, and paired real-world memory evaluation. 
Across the benchmark, 68.9\% of subtasks depend on past observations, making memory a central requirement rather than an optional enhancement.
We also introduced \textbf{PrediMem}, a dual-system memory framework whose predictive coding objective makes hidden states more sensitive to physical state transitions without adding inference-time cost. 


\clearpage


\bibliographystyle{assets/plainnat}
\bibliography{paper}

\clearpage
\newpage
\onecolumn
\beginappendix
\renewcommand{\thefigure}{S\arabic{figure}}
\renewcommand{\thetable}{S\arabic{table}}
\setcounter{figure}{0}
\setcounter{table}{0}

\section{VLM Input Prompt for Data Generation}
\label{sec:appendix-data-generation-prompt}

We use a VLM input prompt template to generate long-horizon robot data by decomposing a coarse task description into executable subtasks and matching each subtask to a predefined planner. Below we show a representative input-only example.

\noindent\textbf{System Prompt.}
\begin{verbatim}
You are a robotic data-generation assistant for long-horizon,
memory-dependent manipulation tasks.

You will be given:
1. One RGB image of the current environment.
2. A coarse long-horizon task description.

Your job is to:
- decompose the task into an ordered list of executable subtasks;
- assign each subtask to one predefined planner from
  {Move, Place, Pour, Open, Close};
- keep the subtasks grounded in the visible scene;
- preserve memory-dependent structure when later steps depend on earlier
  placements, occluded objects, or counted actions;
- return strict JSON only.

Return exactly one field:
- subtasks: a list of ordered subtask entries, each with
  step_id, subtask, planner, and target.
\end{verbatim}

\noindent\textbf{User Prompt.}
\begin{verbatim}
Current environment image:
<image>

Long-horizon task description:
Pour tomato sauce over the cookies and heat them, then pour milk into a cup.

Please decompose this task into executable subtasks and assign the
predefined planner to each step.
\end{verbatim}

\section{VLM Training Prompt and JSON Format}
\label{sec:appendix-json}

We train S2 with a fixed multi-image prompt template. Below we show a representative Task~1 prompt template adapted from \texttt{swift\_compiled\_data.jsonl}.

\noindent\textbf{System Prompt.}
\begin{verbatim}
You are a robotic planning assistant specialized in memory-based task
understanding.

Your task is to infer the current primitive action from multi-image
observation:
1. Historical keyframes from earlier in the same trajectory.
2. A recent 5-timestep visual window ending at the current frame.

Important rules:
- Historical keyframes are earlier than the current window.
- Each timestep contains two images: one agentview_rgb image and one
  eye_in_hand_rgb image.
- The recent 5-timestep window is the primary evidence for current action.
- If there is no keyframe inside the current window, keyframe_positions
  must be an empty list.
- Return strict JSON only. Do not output extra text.
- Return exactly two fields:
  - current_primitive: one primitive from the predefined task1 primitive set.
  - keyframe_positions: 1-indexed keyframe positions inside the recent
    5-timestep window.
\end{verbatim}

\noindent\textbf{User Prompt.}
\begin{verbatim}
Global task: Infer the current primitive action from recent visual history
for a task that stores the cookies and the tomato sauce into the same
target container.

Scene description:
- The manipulation happens on a tabletop.
- The target container is the basket on the right side of the table.
- The square object near the middle is the cookies item.
- The cylindrical object near the middle is the tomato sauce container.
- Each timestep contains two images: one agentview_rgb image and one
  eye_in_hand_rgb image.
- Camera order for every timestep: agentview_rgb, eye_in_hand_rgb.

Current observation:
Recent visual context: 5 consecutive timesteps ending at the current frame
(10 images total; two per timestep, ordered as agentview_rgb followed by
eye_in_hand_rgb):
<image> <image> <image> <image> <image>
<image> <image> <image> <image> <image>

Return strict JSON with fields current_primitive and keyframe_positions
(1-indexed positions inside the recent 5-timestep context).
\end{verbatim}

\noindent\textbf{Assistant Label JSON.}
\begin{verbatim}
{"current_primitive":"pick cookies","keyframe_positions":[]}
\end{verbatim}

\noindent\textbf{Top-Level JSONL Format.}
\begin{verbatim}
{
  "qid": "seed100_order0_win0_r0",
  "messages": [
    {"role": "system", "content": "..."},
    {"role": "user", "content": "..."},
    {"role": "assistant", "content":
      "{\"current_primitive\":\"pick cookies\",\"keyframe_positions\":[]}"}
  ],
  "images": ["data:image/jpeg;base64,...", "... x10"],
  "metadata": {
    "task_id": 1,
    "prompt_style": "breakfast_like",
    "current_primitive": "pick cookies",
    "keyframe_positions": [],
    "camera_keys": ["agentview_rgb", "eye_in_hand_rgb"],
    "history_keyframe_count": 0,
    "num_context_frames": 5,
    "num_context_images": 10,
    "image_size": [256, 256],
    "...": "other window/index fields"
  }
}
\end{verbatim}

\section{Asynchronous Inference Protocol}
\label{sec:appendix-protocol}

Given instruction $\ell$ and observation $o_0$, S2 runs asynchronously on the
recent-frame buffer and memory bank $\mathcal{M}_t$, emitting subtask $c_t$
and keyframe decision $k_t$; the buffered subtask is overwritten whenever a
newer S2 result arrives.
S1 takes the current observation $o_t$ together with the latest subtask $c_t$
at the higher control rate and produces an action chunk
\begin{equation}
\label{eq:vla-action}
a_t = \pi_{\mathrm{S1}}(o_t,\, c_t),
\end{equation}
where $\pi_{\mathrm{S1}}$ denotes the S1 low-level policy (a VLA action head),
$o_t$ is the current visual observation, $c_t$ is the freshest subtask
prediction produced by S2 (overwritten whenever a newer S2 result arrives),
and $a_t$ is the action chunk executed in the next control window.
In our implementation, S2 runs at 1.06\,Hz and S1 at 3.40\,Hz, so each S2
update overlaps roughly 2.92 S1 chunks. Because $\mathcal{M}_t$ retains prior
events, the agent can recover without restarting.
\Cref{tab:async-runtime} shows the detailed runtime profile of \method's
asynchronous inference loop.

\begin{figure}[H]
\centering
\small
\fbox{\parbox{0.92\linewidth}{
\textbf{Algorithm 1:} PrediMem Inference Protocol \\[2pt]
\textbf{Input:} Task instruction $\ell$, initial observation $o_0$ \\
\textbf{Output:} Action sequence $\{a_1, \dots, a_T\}$ \\[4pt]
1: $\mathcal{M}_0 \leftarrow \emptyset$; $g \leftarrow \varnothing$ \\
2: Initialize recent-frame buffer with $o_0$ \\
3: \textbf{for} $t = 1$ to $T$ \textbf{do} \\
4: \quad $a_t \leftarrow \pi_{\mathrm{S1}}(o_t, g)$ \hfill \textit{// High-frequency execution loop} \\
5: \quad \textbf{if} S2 is idle and the recent window is ready \textbf{then} \\
6: \quad\quad Trigger $\mathrm{S2}(\ell, o_t, \mathcal{M}^{rec}_t,\mathcal{M}^{key}_t)$ asynchronously \hfill \textit{// Predict latest subtask} \\
7: \quad \textbf{end if} \\
8: \quad \textbf{if} S2 result $(g_{\mathrm{new}}, k_{\tau})$ is available \textbf{then} \\
9: \quad\quad $g \leftarrow g_{\mathrm{new}}$ \hfill \textit{// Refresh current subtask} \\
10: \quad\quad \textbf{if} $k_{\tau} = 1$ \textbf{then} $\mathcal{M}_t^{\mathrm{key}} \leftarrow \mathcal{M}_t^{\mathrm{key}} \cup \{o_{\tau}\}$ \textbf{end if} \hfill \textit{// Memory write} \\
11: \quad \textbf{end if} \\
12: \quad $\mathcal{M}_t^{\mathrm{rec}} \leftarrow \text{last } W \text{ frames}$ \hfill \textit{// Update recent sliding window} \\
13: \textbf{end for}
}}
\caption{\textbf{Inference protocol of PrediMem.} S2 asynchronously predicts the latest subtask and keyframe decisions from recent visual history, while S1 executes precise actions at a high control frequency under the freshest available subtask.}
\label{alg:inference}
\end{figure}

\begin{table}[!h]
\centering
\caption{\textbf{Runtime profile of the asynchronous inference loop.}
Runtime measurements of our current asynchronous implementation, denoted \texttt{a0005}.}
\label{tab:async-runtime}
\small
\setlength{\tabcolsep}{4pt}
\renewcommand{\arraystretch}{1.08}
\resizebox{\linewidth}{!}{%
\begin{tabular}{ll}
\toprule
\textbf{Metric} & \textbf{PrediMem async implementation (\texttt{a0005})} \\
\midrule
High-level VLM refresh rate & 1.06 Hz steady state (p50 0.939 s) \\
High-level VLM latency & p50 0.939 s; p95 1.136 s; mean 1.752 s including cold start \\
Low-level VLA frequency & 3.40 Hz (mean 0.294 s; p50 0.289 s; p95 0.365 s) \\
VLM:VLA scheduling & one VLM update spans $\sim$2.92 VLA chunks \\
\bottomrule
\end{tabular}
}
\end{table}

\section{Memory-Dependent Subtask Ratio Annotation}
\label{sec:appendix-memory-ratio}
We define a subtask as \emph{memory-dependent} when the correct current subtask cannot be determined from the current observation alone and requires more information from earlier subtasks or observations.
Equivalently, a subtask is counted as memory-dependent if removing the execution history or observation would make the correct high-level decision ambiguous.
For task $i$ with $n_i$ subtasks and $m_i$ memory-dependent subtasks, we compute the task-level ratio as
\begin{equation}
r_i = \frac{m_i}{n_i}.
\end{equation}
The benchmark-level memory-dependent subtask ratio is computed over all subtasks:
\begin{equation}
R_{\mathrm{mem}} =
\frac{\sum_i m_i}{\sum_i n_i}.
\end{equation}
For RoboMemArena, this gives $R_{\mathrm{mem}}=104/151=68.9\%$, where 104 denotes that total number of memory-dependent subtasks
across all tasks, and 151 denotes the total number of subtasks by summing all tasks.

Our definition encompasses these \textbf{four common forms} of memory demand, but is not limited to them.
\textbf{Occlusion} requires remembering the location of a target object after it becomes hidden inside a container, for example, when the robot must place another object into the same container.
\textbf{Counting} requires remembering how many times an action has already been performed, such as whether it has already been poured once and needs to be poured again.
\textbf{Transferring} requires remembering the source--target mapping between containers that are visually similar.
\textbf{Sequence} requires remembering which prerequisite subtasks have already been completed before executing the next step.
For example, in a drawer-search task, the first drawer opening may not require memory, whereas later subtasks should depend on which drawers have already been inspected and what was observed in each drawer. In this case, if 7 out of 9 subtasks require such history-dependent information, the memory ratio is $7/9$.

For benchmarks with a small number of tasks, we manually inspect task descriptions, keyframe annotation and subtask decompositions.
For larger benchmarks, we use an LLM-assisted first pass with a fixed rubric and then manually check the ambiguous cases.
The simplified rubric is:
\begin{verbatim}
We classify each subtask as memory-dependent or memory-free according to
whether it requires information from earlier observations or task states.

A subtask is memory-free if the required object and target state are visible
in the current observation, e.g., picking up a visible cup or placing a 
visible cup on the table.

A subtask is memory-dependent if its execution depends on earlier task
state or observation,including but not limited to:
1. Occlusion: Remembering that an object was placed into an occlusive container.
2. Counting: Remembering how many times an action has already been performed.
3. Transferring: Remembering the source-target mapping between similar containers.
4. Sequence: Remembering which prerequisite subtasks have already been completed.

\end{verbatim}

\section{Benchmark Task Details}
\label{sec:appendix-benchmark}

{
\small
\setlength{\tabcolsep}{4pt}
\renewcommand{\arraystretch}{1.05}
\setlength{\emergencystretch}{2em}

\begin{longtable}{
>{\raggedright\arraybackslash}p{2.4cm}
>{\centering\arraybackslash}p{0.7cm}
>{\centering\arraybackslash}p{1.1cm}
>{\raggedright\arraybackslash}p{2.0cm}
>{\raggedright\arraybackslash}p{4.9cm}
}
\caption{\textbf{Benchmark task descriptions.} Overview of all 26 RoboMemArena tasks, their corresponding memory types, average total timesteps, and key challenges.}
\label{tab:appendix-benchmark-tasks} \\
\toprule
\textbf{Task Name} & \textbf{Memory Type} & \textbf{Avg. \#Steps} & \textbf{Task Challenge} & \textbf{Brief Description} \\
\midrule
\endfirsthead

\caption[]{\textbf{Benchmark task descriptions.} (Continued)} \\
\toprule
\textbf{Task Name} & \textbf{Memory Type} & \textbf{Avg. \#Steps} & \textbf{Task Challenge} & \textbf{Brief Description} \\
\midrule
\endhead

\midrule
\multicolumn{5}{r}{\textit{Continued on next page...}} \\
\endfoot

\bottomrule
\endlastfoot

\multicolumn{5}{l}{\texttt{Task Suite: Multi-Object Transferring}} \\
Transfer Chocolate Butter & T & 866 & transferring & Pick and place chocolate and butter from plate1 to plate2, respectively. \\
\rowcolor{gray!8}
Transfer Butter Cheese & T & 779 & transferring & Pick and place butter and cheese from plate1 to plate2, respectively. \\
Transfer Popcorn Cookies & T & 779 & transferring & Pick and place popcorn and cookies from plate1 to plate2, respectively. \\
\rowcolor{gray!8}
Transfer Sauce Milk Juice & T & 1265 & transferring & Pick and place tomato sauce, milk, and orange juice from cabinet1 to cabinet2. \\
\midrule

\multicolumn{5}{l}{\texttt{Task Suite: Multi-Object Occlusion}} \\
Put Butter in Not-Empty Drawer & O & 1020 & occlusion & Open all drawers in order. Put butter into the drawer that already contains an object. \\
\rowcolor{gray!8}
Put Butter in Empty Drawer & O & 1806 & occlusion & Open all drawers in order. Put butter into the empty drawer. \\
Put Cookies Butter into Drawer Respectively & O + C & 1835 & occlusion, multi-placement & Open all drawers in order. Put cookies into the top drawer and put butter into another drawer. \\
\rowcolor{gray!8}
Put Cookies Chocolate into Middle Drawer & O & 1370 & occlusion & Open all drawers in order. Put cookies into the middle drawer and then put chocolate into the same drawer. \\
Put Butter Cookies into Middle Drawer & O & 1377 & occlusion & Open all drawers in order. Put butter into the middle drawer and then put cookies into the same drawer. \\
\rowcolor{gray!8}
Put Cookies Chocolate into Drawer Respectively & O & 1832 & occlusion, multi-placement & Open all drawers in order. Put cookies into the top drawer and put chocolate into another drawer. \\
Put Butter Chocolate into Middle Drawer & O & 1502 & occlusion & Open all drawers in order. Put butter into the middle drawer and then put chocolate into the same drawer. \\
\rowcolor{gray!8}
Put Cookies Chocolate into Microwave & O & 1195 & occlusion & Put cookies into the microwave and then put chocolate into the location where the cookies were placed. \\
Put Butter Chocolate into Microwave & O & 1175 & occlusion & Put butter into the microwave and then put chocolate into the location where the butter was placed. \\
\rowcolor{gray!8}
Put Cream Popcorn into Microwave & O & 1175 & occlusion & Put cream into the microwave and then put popcorn into the location where the cream was placed. \\
Put Cookies Popcorn into Microwave & O & 1195 & occlusion & Put cookies into the microwave and then put popcorn into the location where the cookies were placed. \\
\midrule

\multicolumn{5}{l}{\texttt{Task Suite: Multi-Object Counting}} \\
Pour Sauce on Cookies $\times$2 Place Sauce into Drainer & C + O & 624 & counting, occlusion & Pour tomato sauce over cookies twice and place the sauce bottle into the bowl drainer. \\
\rowcolor{gray!8}
Pour Sauce on Frypan $\times$2 Place Sauce into Drainer & C + O & 537 & counting, occlusion & Pour tomato sauce over the frypan twice and place the sauce bottle into the bowl drainer. \\
Pour Sauce Twice over Chocolate in Frypan Place Sauce into Drainer & C & 910 & counting & Pick and place chocolate into the frypan, pour tomato sauce over it twice, then place the sauce bottle into the bowl drainer. \\
\rowcolor{gray!8}
Pour Sauce $\times$2 over Butter in Frypan & C & 958 & counting & Put butter into the frypan and pour sauce over it twice. \\
Pour Wine into Mug Twice & C & 472 & counting & Pour wine into the mug twice. \\
\rowcolor{gray!8}
Pour Milk Twice over Butter in Frypan & C & 1055 & counting & Pick and place butter into the frypan, then pour milk over it twice. \\
Pour Milk $\times$2 over Mug Place Milk into Drainer & C + O & 594 & counting, occlusion & Pick milk from the table, pour it into the mug twice, then place the milk container into the bowl drainer. \\
\midrule

\multicolumn{5}{l}{\texttt{Task Suite: Multi-Object Sequence}} \\
Put Cookies Sauce into Basket in Order & S & 742 & sequence & Pick and place cookies into the basket, then pick and place tomato sauce into the same basket. \\
\rowcolor{gray!8}
Put Butter Popcorn into Basket in Order & S & 708 & sequence & Pick and place butter into the basket, then pick and place popcorn into the same basket. \\
Put Cream Chocolate into Basket in Order & S & 708 & sequence & Pick and place cream into the basket, then pick and place chocolate into the same basket. \\
\rowcolor{gray!8}
Pour Sauce $\times$2 Put Cookies into Microwave & S & 1565 & sequence & Pour tomato sauce over cookies twice, then put the cookies into the microwave. \\
\end{longtable}
}

\section{Reactive Policy Failure Modes}
\label{sec:appendix-failures}

The quantitative gap between reactive and memory-augmented policies maps to two concrete failures.
First, the reactive policy cannot distinguish whether a drawer has already been checked once the visual state resets.
Second, it cannot preserve the instruction-level constraint that all drawers must be opened before the final placement.
Detailed failure modes are provided in \Cref{tab:case-analysis}.
\begin{table}[H]
\caption{\textbf{Failure modes of the reactive $\pi_{0.5}$ baseline on a memory-dependent drawer task.}}
\label{tab:case-analysis}
\centering
\small
\begin{tabular}{lp{0.62\columnwidth}}
\toprule
\textbf{Failure mode} & \textbf{Description} \\
\midrule
State aliasing & After the first drawer is closed, the observation becomes visually similar to the initial frame. Without memory, the policy repeatedly opens the same drawer and enters a loop. \\
Constraint forgetting & After finding one locally valid empty drawer, the policy places the butter immediately and forgets the global instruction to inspect all drawers before deciding. \\
\bottomrule
\end{tabular}
\end{table}

\section{Real-World Task Details}
\label{sec:appendix-realworld}

{
\small
\begin{longtable}{
>{\raggedright\arraybackslash}p{2.7cm}
>{\centering\arraybackslash}p{0.9cm}
>{\centering\arraybackslash}p{1.2cm}
>{\raggedright\arraybackslash}p{2.4cm}
>{\raggedright\arraybackslash}p{5.2cm}
}
\caption{\textbf{Real-world task descriptions.} Overview of the five physical-robot tasks, their corresponding memory types, average total timesteps, and key challenges.}
\label{tab:appendix-realworld-tasks} \\
\toprule
\textbf{Task Name} & \textbf{Memory Type} & \textbf{Avg. \#Steps} & \textbf{Task Challenge} & \textbf{Brief Description} \\
\midrule
\endfirsthead

\caption[]{\textbf{Real-world task descriptions.} (Continued)} \\
\toprule
\textbf{Task Name} & \textbf{Memory Type} & \textbf{Avg. \#Steps} & \textbf{Task Challenge} & \textbf{Brief Description} \\
\midrule
\endhead

\midrule
\multicolumn{5}{r}{\textit{Continued on next page...}} \\
\endfoot

\bottomrule
\endlastfoot

\multicolumn{5}{l}{\texttt{Task Suite: Real-World Evaluation}} \\
Pour Bottle $\times$2 & C & 866 & counting & Pour water from the bottle into the cup twice. \\
\rowcolor{gray!8}
Brush Plates with Swap & C + S & 779 & sequence & Brush three plates in order. \\
Transfer Objects & T & 779 & transferring & Transfer the watermelon and the carrot from one plate to another. \\
\rowcolor{gray!8}
Shell Game & O + S & 779 & occlusion, tracking & Hide the target under one cup, swap the positions of the three cups, and identify the cup containing the target. \\
Breakfast from Human & C + S & 1265 & imitation, sequence & A human demonstrates how to make breakfast, and the robot imitates the demonstrated breakfast-making sequence. \\
\end{longtable}
}

\begin{figure*}[!h]
    \centering
    \includegraphics[width=\textwidth]{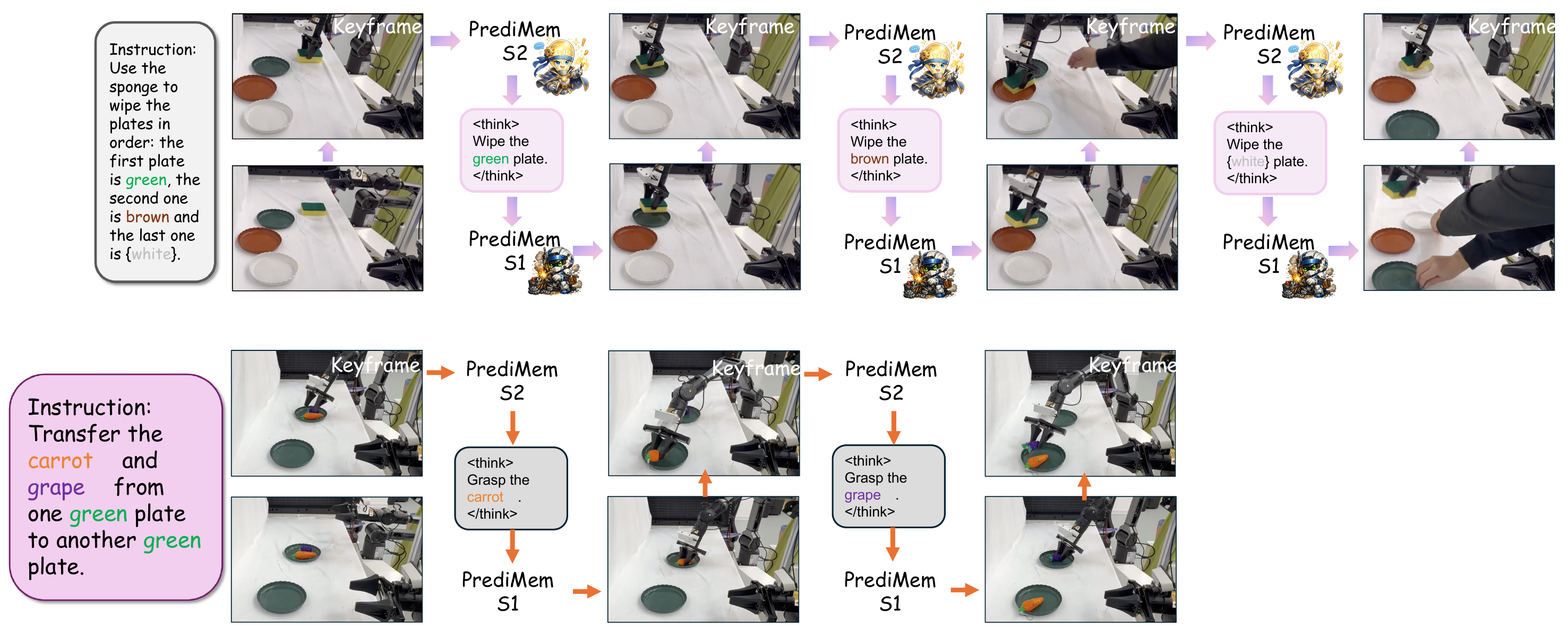}
    \caption{\textbf{Real-world task demonstrations.}
    Representative snapshots from our physical-robot evaluation suite.
    The figure summarizes the real-world task settings, example execution frames, and the dual-arm platform layout in a format similar to standard real-robot evaluation overviews.}
    \label{fig:real-world-demo}
\end{figure*}

\section{Verification-Step Distribution}
\label{sec:appendix-verification}

\begin{figure}[H]
    \centering
    \includegraphics[width=\linewidth]{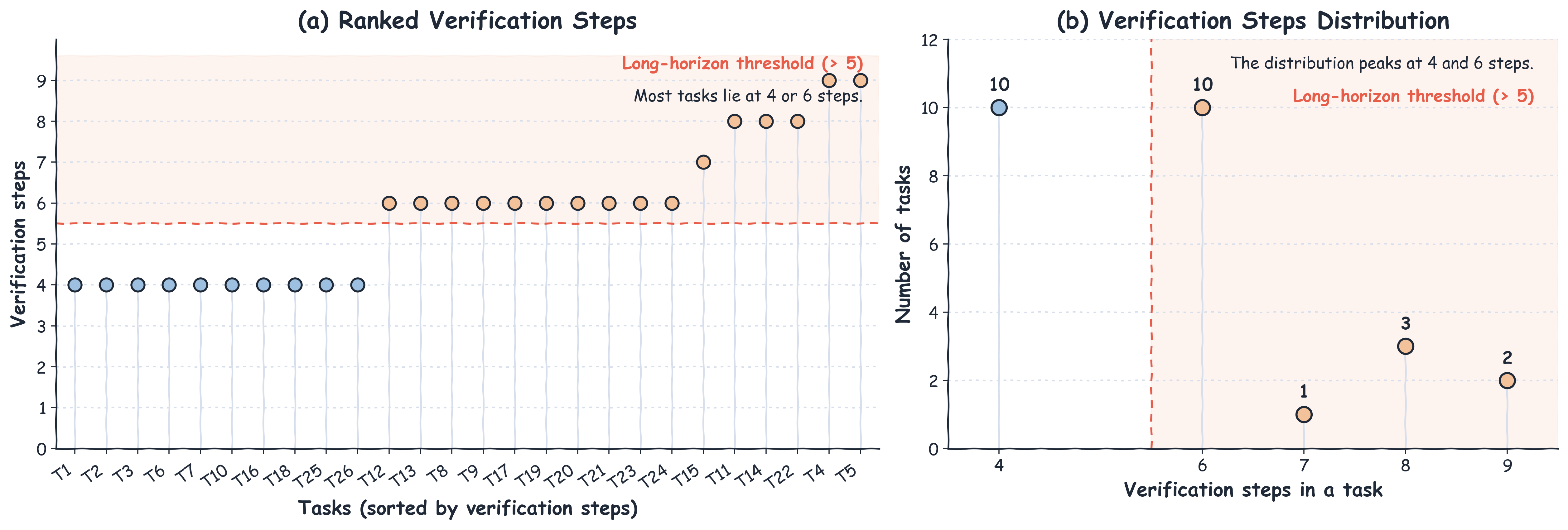}
    \caption{\textbf{Distribution of verification steps per task.}
    \textbf{(a)} Each task contains 3--9 verification steps. Most exceed the long-horizon threshold of five steps.
    \textbf{(b)} Overall histogram of verification step counts.
    The distribution ensures that the cumulative success rate (CSR) metric has sufficient resolution to distinguish agents with differing levels of memory capability.}
    \label{fig:verification}
\end{figure}

\clearpage

\end{document}